\documentclass[11pt,draftcls,onecolumn]{IEEEtran}

\usepackage{graphicx,bm}

\usepackage[ruled,vlined]{algorithm2e}

\renewcommand{\(}{\left(}
\renewcommand{\)}{\right)}
\renewcommand{\[}{\left[}
\renewcommand{\]}{\right]}
\renewcommand{\c}{\mathbf{c}}

\renewcommand{\S}{\mathbf{S}}

\renewcommand{\H}{\mathbf{H}}

\newcommand{\0}{\mathbf{0}}
\newcommand{\D}{\mathbf{D}}

\newcommand{\x}{\mathbf{x}}
\newcommand{\I}{\mathbf{I}}

\newcommand{\C}{\mathbf{C}}

\newcommand{\A}{\mathbf{A}}

\newcommand{\U}{\mathbf{U}}
\newcommand{\M}{\mathbf{M}}

\renewcommand{\u}{\mathbf{u}}
\renewcommand{\v}{\mathbf{v}}
\newcommand{\Q}{\mathbf{Q}}
\newcommand{\K}{\mathbf{K}}

\newcommand{\X}{\mathbf{X}}

\newcommand{\B}{\mathbf{B}}

\renewcommand{\log}[1]{{\rm{log}}#1}

\newtheorem{lemma}{Lemma}
\newtheorem{proposition}{Proposition}

\title{Decomposable Principal Component Analysis}

\author{Ami~Wiesel and Alfred~O.~Hero~III\\ Department of Electrical Engineering and Computer Science\\ University of Michigan, Ann Arbor, MI 48109, USA\\ E-mails: \{amiw,hero\}@umich.edu }

\begin{document}

\maketitle

\begin{abstract}
  We consider principal component analysis (PCA) in decomposable Gaussian graphical models. We exploit the prior information in these models in order to distribute its computation. For this purpose, we reformulate the problem in the sparse inverse covariance (concentration) domain and solve the global eigenvalue problem using a sequence of local eigenvalue problems in each of the cliques of the decomposable graph. We demonstrate the application of our methodology in the context of decentralized anomaly detection in the Abilene backbone network. Based on the topology of the network, we propose an approximate statistical graphical model and distribute the computation of PCA.
\end{abstract}

\section{Introduction}

We consider principal component analysis (PCA) in Gaussian graphical
models. PCA is a classical dimensionality reduction method which is frequently used in statistics and machine learning \cite{hastietf01,Anderson:book}. The first principal components of a multivariate are its orthogonal linear combinations which preserve most of the variance. In the Gaussian case, PCA has special properties which make it especially favorable: it is the best linear approximation of the data and it provides independent components. On the other hand, Gaussian graphical models, also known as covariance selection models, provide a graphical representation of the conditional independence structure within the Gaussian distribution \cite{lauritzen:book,dempster:72}. Exploiting the extensive knowledge and literature on graph theory, graphical models allow for efficient distributed implementation of statistical inference algorithms, e.g., the well known belief propagation method and the junction tree algorithm \cite{YairWeiss10012001,jordan:book}. In particular, decomposable graphs, also known as chordal or triangulated graphs, provide simple and intuitive inference methods due to their appealing structure. Our main contribution is the application of decomposable graphical models to PCA which we nickname DPCA, where D denotes both {\em{Decomposable}} and {\em{Distributed}}.

The main motivation for distributed PCA is decentralized dimensionality reduction. It plays a leading role in distributed estimation and compression theory in wireless sensor networks \cite{zhu:may2005,Schizas:2007,xiao:2006,Gastpar:06,roy:08}, and decentralized data mining techniques \cite{Kargupta:2001,bai:book2005,qu:icdm2002}. It is also used in anomaly detection in computer networks \cite{Lakhina:2004,Chhabra:2008,huang:nips2006}. In particular, \cite{Gastpar:06,roy:08} proposed to approximate the global PCA using a sequence of conditional local PCA solutions. Alternatively, an approximate solution which allows a tradeoff between performance and communication requirements was proposed in \cite{huang:nips2006} using eigenvalue perturbation theory.

DPCA is an efficient implementation of distributed PCA based on a prior graphical model. Unlike the above references it does not try to approximate PCA, but yields an exact solution up to on any given tolerance. On the other hand, it assumes additional prior knowledge in the form of a graphical model which previous works did not take into account. Although, it is interesting to note that the Gauss Markov source example in \cite{Gastpar:06,roy:08} is probably the most celebrated decomposable graphical model. Therefore, we now address the availability of such prior information. In general, practical applications do not necessarily satisfy any obvious conditional independence structure. In such scenarios, DPCA can be interpreted as an approximate PCA method that allows a tradeoff between accuracy and decentralization by introducing sparsity. In other problems it is reasonable to assume that an unknown structure exists and can be learned from the observed data using existing methods such as \cite{banerjee-2007,friedman-2007,yuan:2007}. Alternatively, a graphical model can be derived from non-statistical prior knowledge on the specific application. An intuitive example is distributed networks in which the topology of the network suggests a statistical graph as exploited in \cite{willsky:SPM}. Finally, we emphasize that even if a prior graphical model is available, it does not necessarily satisfy a decomposable form. In this case, a decomposable approximation can be obtained using classical graph theory algorithms \cite{jordan:book}.

PCA can be interpreted as maximum likelihood (ML) estimation of the covariance using the available data followed by its eigenvalue decomposition. When a prior graphical model is available, PCA can still be easily obtained by adjusting the ML estimation phase to incorporate the prior conditional independence structure using existing methods \cite{lauritzen:book,dempster:72}, and then computing the eigenvalue decomposition (EVD). The drawback to this approach is that it does not exploit the structure of the graph in the EVD phase. This disadvantage is the primary motivation to DPCA which is specifically designed to utilize the structure of Gaussian graphical models. Decomposable covariance selection models result in sparse concentration (inverse covariance) matrices which can be estimated in a decentralized manner. Therefore, we propose to reformulate DPCA in the concentration domain and solve the global EVD using a sequence of local EVD problems in each of the cliques of the decomposable graph with a small amount of message passing. This allows for distributed implementation according to the topology of the graph and reduces the need to collect all the observed data in a centralized processing unit. When the algorithm terminates, each clique obtains its own local version of the principal components.

To illustrate DPCA we apply it to distributed anomaly detection in computer networks \cite{Lakhina:2004,huang:nips2006}. In this context, DPCA learns a low dimensional model of the normal traffic behavior and allows for simple outlier detection. This application is natural since the network's topology provides a physical basis for constructing an approximate a graphical model.  For example, consider two nodes which are geographically distant and linked only through a long path of nodes. It is reasonable to believe that these two sensors are independent conditioned on the path, but a theoretical justification of this assertion is difficult and depends on the specific problem formulation. We examine the validity of this claim in the context of anomaly detection in the Abilene network using a real-world dataset. We propose an approximate decomposition of the Abilene network, enable the use of DPCA and obtain a fully distributed anomaly detection method.

The outline of the paper is as follows. Decomposable graphs are easy to explain using a special graph of two cliques which is their main building block. Therefore, we begin in section \ref{sec_2dpca} by introducing the problem formulation and solution to DPCA in this simple case. The generalization to decomposable graphs is presented in section \ref{sec_dpca} which consists of their technical definitions followed by a recursive application of the two cliques solution. We demonstrate the use of DPCA using two numerical examples. First, in Section \ref{sec_num} we simulate our proposed algorithm in a synthetic tracking scenario. Second, in Section \ref{sec_abilene} we illustrate its application to anomaly detection using a real-world dataset from the Abilene backbone network. Finally, in Section \ref{sec_conc} we provide concluding remarks and address future work.

The following notation is used. Boldface upper case letters denote
matrices, boldface lower case letters denote column vectors, and
standard lower case letters denote scalars. The superscripts
$(\cdot)^T$ and $(\cdot)^{-1}$ denote the transpose and matrix inverse, respectively. The cardinality of a set $a$ is denoted by $|a|$.  The matrix $\I$ denotes the identity,
${\rm{eig}}_{\min}\(\X\)$ is the minimum eigenvalue of square symmetric matrix $\X$, $\u_{\rm{null}}\(\X\)$ is a null vector of $\X$, ${\rm{eig}}_{\max}\(\X\)$ is the maximum eigenvalue of $\X$, and
$\X\succ\0$ means that $\X$ is positive definite. Finally, we use indices in the subscript $\[\x\]_a$ or $\[\X\]_{a,b}$ to denote sub-vectors or sub-matrices, respectively, and $\[\X\]_{a,:}$ denotes the sub-matrix formed by the $a$'th rows in $\X$. Where possible, we omit the brackets and use $\x_a$ or $\X_{a,b}$ instead.

\section{Two clique DPCA}\label{sec_2dpca}
In this section, we introduce DPCA for a simple case which will be the building block for the general algorithm.

\begin{figure}\center
\includegraphics[width=0.40\textwidth]{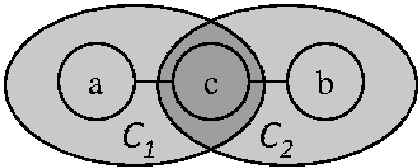}
\caption{Graphical model with two cliques modeling a 3 node network in which $a$ and $b$ are conditionally independent given $c$.
\label{fig_twoclique}}
\end{figure}

\subsection{Problem Formulation}
Let $\x=\[\x_a^T\;\x_c^T\;\x_b^T\]^T$ be a length $p$, zero mean Gaussian random vector in which $\[\x\]_a$ and $\[\x\]_b$ are independent conditionally on $\[\x\]_c$ where $a$, $c$ and $b$ are disjoint subsets of indices. For later use, we use graph terminology and define two cliques of indices $C_1=\{a,c\}$ and $C_2=\{c,b\}$ coupled through the separator $S=\{c\}$ (see Fig. \ref{fig_twoclique}). We assume that the covariance matrix of $\x$ is unknown, but the conditional independence structure (defined through index sets $C_1$ and $C_2$) is known.

The input to DPCA is a set of $n$ independent and identically distributed realizations of $\x$ denoted by $\x_i$ for $i=1,\cdots,n$. More specifically, this input is distributed in the sense that the first clique has access to $\[\x_i\]_{C_1}$ for $i=1,\cdots,n$, whereas the second clique has access only to $\[\x_i\]_{C_2}$ for $i=1,\cdots,n$. Using this data and minimal message passing between the two cliques, DPCA searches for the linear combination $X=\u^T\x$ having maximal variance. When the algorithm terminates, each of the cliques obtains its own local version of $\u$, i.e., the sub-vectors $\[\u\]_{C_1}$ and $\[\u\]_{C_2}$.

The following subsections present the proposed solution to DPCA. It involves two main stages: covariance estimation and principal components computation.

\subsection{Solution: covariance matrix estimation}
First, the covariance matrix of $\x$ is estimated using the maximum likelihood (ML) technique. Due to the known conditional independence structure, the ML estimate has a simple closed form solution which can be computed in a distributed manner (more details about this procedure can be found in \cite{lauritzen:book}). Each clique and the separator computes their own local sample covariance matrices
\begin{eqnarray}
  \tilde\S^{C_1,C_1}&=&\frac{1}{n}\sum_{i=1}^n\[\x_i\]_{C_1}\[\x_i\]_{C_1}^T\\
  \tilde\S^{C_2,C_2}&=&\frac{1}{n}\sum_{i=1}^n\[\x_i\]_{C_2}\[\x_i\]_{C_2}^T\\
  \tilde\S^{S,S}&=&\frac{1}{n}\sum_{i=1}^n\[\x_i\]_{S}\[\x_i\]_{S}^T,
\end{eqnarray}
where the tilde and the superscripts are used to emphasize that these are local estimates. Similarly, the local concentration matrices, also known as precision matrices, are defined as
\begin{eqnarray}
  \tilde\K^{C_1,C_1}&=&\(\tilde\S^{C_1,C_1}\)^{-1}\\
  \tilde\K^{C_2,C_2}&=&\(\tilde\S^{C_2,C_2}\)^{-1}\\
  \tilde\K^{S,S}&=&\(\tilde\S^{S,S}\)^{-1},
\end{eqnarray}
where it is assumed that the matrices are non-singular (otherwise, the ML estimate does not exist). Next, the global ML concentration matrix $\K$ is obtained by requiring
\begin{eqnarray}\label{Kab}
  \K_{a,b}=\K^T_{b,a}=\0
\end{eqnarray}
due to the conditional independence of $\x_a$ and $\x_b$ given $\x_c$. The global solution is
\begin{eqnarray}\label{Kabc}
  \K=\[\begin{array}{cc}
         \tilde\K^{C_1,C_1} & \begin{array}{c}
                                \0 \\
                                \0
                              \end{array}
          \\
         \begin{array}{cc}
           \0 & \0
         \end{array}
          & \0
       \end{array}
  \]+\[\begin{array}{cc}
         \0 & \begin{array}{cc}
               \0 & \0
             \end{array}
          \\
         \begin{array}{c}
           \0 \\
           \0
         \end{array}
          & \tilde\K^{C_2,C_2}
       \end{array}
  \]-\[\begin{array}{ccc}
         \0 & \0 & \0 \\
         \0 & \tilde\K^{S,S} & \0 \\
         \0 & \0 & \0
       \end{array}
  \].
\end{eqnarray}
It is easy to see that the sub-matrices associated with the cliques are perturbations of to their local versions:
\begin{eqnarray}
\label{Ksubblocks1}  \[\K\]_{C_1,C_1}&=&\tilde\K^{C_1,C_1}+\[\begin{array}{cc}
                                               \0 & \0\\
                                               \0 & \M_{b}
                                             \end{array}
  \]\\
\label{Ksubblocks2}  \[\K\]_{C_2,C_2}&=&\tilde\K^{C_2,C_2}+\[\begin{array}{cc}
                                               \M_{a} & \0 \\
                                               \0  & \0
                                             \end{array}\]
\end{eqnarray}
and require only message passing via $\M_{b}$ and $\M_{a}$:
\begin{eqnarray}
  \M_{b}&=&\[\tilde\K^{C_2,C_2}\]_{S,S}-\tilde\K^{S,S}\\
  \M_{a}&=&\[\tilde\K^{C_1,C_1}\]_{S,S}-\tilde\K^{S,S}.
\end{eqnarray}
The dimension of these messages is equal to $|S|$ which is presumably small. Thus, the global ML concentration matrix can be easily found in a distributed manner.

The global covariance estimate is simply defined as the inverse of its concentration $\S=\K^{-1}$. It is consistent with the local estimates of its sub-matrices:
\begin{eqnarray}
  \[\S\]_{C_1,C_1}&=&\tilde\S^{C_1,C_2}\\
  \[\S\]_{C_2,C_2}&=&\tilde\S^{C_2,C_2},
\end{eqnarray}
but there is no special intuition regarding its $\[\S\]_{a,b}$ and $\[\S\]_{b,a}$ sub-blocks.

\subsection{Solution: first principal eigenvalue}\label{sec_eigval}
Given the global ML covariance estimate $\S$, the PCA objective function is estimated as
\begin{eqnarray}
 \u^T\S\u,
\end{eqnarray}
which is maximized subject to a norm constraint to yield
\begin{eqnarray}\label{cpca}
 {\rm{eig}}_{\max}\(\S\)=\left\{\begin{array}{ll}
  \max_\u & \u^T\S\u \\
  {\rm{s.t.}} & \u^T\u=1.
 \end{array}\right.
 \end{eqnarray}
This optimization gives both the maximal eigenvalue of $\S$ and the its eigenvector $\u$.

The drawback to the above solution is that the EVD computation requires centralized processing and does not exploit the structure of $\K$. Each clique needs to send its local covariance to a central processing unit which constructs $\S$ and computes its maximal eigenvalue and eigenvector. We will now provide an alternative distributed DPCA algorithm in which each clique uses only local information along with minimal message passing in order to calculate its local version of ${\rm{eig}}_{\max}\(\S\)$ and $\u$.

Our first observation is that DPCA can be equivalently solved in the
concentration domain instead of the covariance
domain. Indeed, it is well known that
\begin{eqnarray}
 {\rm{eig}}_{\max}\(\S\)=\frac{1}{{\rm{eig}}_{\min}\(\K\)},
\end{eqnarray}
when the inverse $\K=\S^{-1}$ exists. The corresponding eigenvectors are also identical. The advantage of working with $\K$ instead of $\S$ is that we can directly exploit $\K$'s sparsity as expressed in (\ref{Kab}).

Before continuing it is important to address the question of
singularity of $\S$. One may claim that working in the
concentration domain is problematic since $\S$ may be
singular. This is indeed true but is not a critical disadvantage since graphical models allow for well conditioned estimates under small sample sizes. For example, classical ML exists only if $n\geq p$, whereas the ML described above requires the less stringent condition $n\geq\max\{|C_1|,|C_2|\}$ \cite{lauritzen:book}. In fact, the ML covariance is defined as the inverse of its concentration, and thus the issue of singularity is an intrinsic problem of ML estimation rather
than the DPCA solution.

We now return to the problem of finding
\begin{eqnarray}
  \lambda={\rm{eig}}_{\min}\(\K\)
\end{eqnarray}
in a distributed manner. We begin by expressing $\lambda$ as a trivial line-search problem:
\begin{eqnarray}\label{eigasopt}
  \lambda = \sup \quad t \quad {\rm{s.t.}} \quad t<{\rm{eig}}_{\min}\(\K\)
\end{eqnarray}
and note that the objective is linear and the constraint set is convex. It can be solved using any standard line-search algorithm, e.g. bisection. At first, this representation seems useless as we still need to evaluate ${\rm{eig}}_{\min}\(\K\)$ which was our original goal. However, the following proposition shows that checking the feasibility of a given $t$ can be done in a distributed manner.
\begin{proposition}
  Let $\K$ be a symmetric matrix with $\K_{a,b}=\K_{b,a}^T=\0$. Then, the constraint
  \begin{eqnarray}\label{eigasopt1}
    t<{\rm{eig}}_{\min}\(\[\begin{array}{ccc}
    \K_{a,a} & \K_{a,c} & \0 \\
    \K_{c,a} & \K_{c,c} & \K_{c,b} \\
    \0 & \K_{b,c} & \K_{b,b}
  \end{array}
 \]\)
  \end{eqnarray}
  is equivalent to the following pair of constraints
  \begin{eqnarray}
\label{prop2}    t&<&{\rm{eig}}_{\min}\(\K_{b,b}\)\\
\label{prop1}    t&<&{\rm{eig}}_{\min}\(\K_{C_1,C_1}-\[\begin{array}{cc}
                                               \0 & \0\\
                                               \0 & \M\(t\)
                                             \end{array}
  \]\)
  \end{eqnarray}
  with the {\em{message matrix}} defined as
  \begin{eqnarray}
    \label{message}\M\(t\)=\K_{c,b}\(\K_{b,b}-t\I\)^{-1}\K_{b,c}.
  \end{eqnarray}
\end{proposition}
\begin{IEEEproof}
  The proof is obtained by rewriting (\ref{eigasopt1}) as a linear matrix inequality
  \begin{eqnarray}\label{eig2lmi}
    \[\begin{array}{ccc}
    \K_{a,a} & \K_{a,c} & \0 \\
    \K_{c,a} & \K_{c,c} & \K_{c,b} \\
    \0 & \K_{b,c} & \K_{b,b}
  \end{array}
 \]-t\I\succ\0
  \end{eqnarray}
  and decoupling this inequality using the following lemma:
  \begin{lemma}[Schur's Lemma {\cite[Appendix A5.5]{boyd:2003}}]\label{schur} Let $\X$ be a symmetric matrix partitioned as
  \begin{eqnarray}
  \X=\[\begin{array}{cc}
  \A & \B \\
  \B^T & \C
\end{array}\].
\end{eqnarray}
Then, $\X\succ\0$ if and only if $\A\succ \0$ and $\C-\B^T\A^{-1}\B\succ \0$.
\end{lemma}
  Applying Schur's Lemma to (\ref{eig2lmi}) with $\A=\K_{C_1,C_1}$ and rearranging yields
\begin{eqnarray}
\label{tIKbb}    t\I&\prec&\K_{b,b}\\
\label{tIKac}    t\I&\prec&\K_{C_1,C_1}-\[\begin{array}{cc}
                                               \0 & \0\\
                                               \0 & \M\(t\)
                                             \end{array}
  \].
\end{eqnarray}
Finally, (\ref{prop2}) and (\ref{prop1}) are obtained by rewriting (\ref{tIKbb}) and (\ref{tIKac}) as eigenvalue inequalities, respectively.
\end{IEEEproof}

Proposition 1 provides an intuitive distributed solution to (\ref{eigasopt}). For any given $t$ we can check the feasibility by solving local eigenvalue problems and message passing via $\M\(t\)$ whose dimension is equal to the cardinality of the separator. The optimal global eigenvalue is then defined as the maximal globally feasible $t$ .

We note that the solution in Proposition 1 is asymmetric with respect to the cliques. The global constraint is replaced by two local constraints regarding clique $C_1=\{a,c\}$ and the remainder $\{b\}$. Alternatively, we can exchange the order and partition the indices into $\{a\}$ and $C_2=\{c,b\}$. This asymmetry will become important in the next section when we extend the results to general decomposable graphs.

\subsection{Solution: first principal eigenvector}\label{sec_eigvec}
After we obtain the minimal eigenvalue $\lambda$, we can easily recover its corresponding eigenvector $\u$. For this purpose, we define $\Q=\K-\lambda\I$ and obtain $\u=\u_{\rm{null}}\(\Q\)$. The matrix $\Q$ follows the same block sparse structure as $\K$, and the linear set of equations $\Q\u=\0$ can be solved in a distributed manner. There are two possible solutions. Usually, $\Q_{bb}$ is non-singular in which case
the solution is
\begin{eqnarray}\label{uac}
    \[\u\]_{C_1}&=&\u_{\rm{null}}\(\Q_{C_1,C_1}-\[\begin{array}{cc}
                                               \0 & \0\\
                                               \0 & \M
                                             \end{array}
  \]\)\\
\label{ub}  \[\u\]_b&=&-\Q_{b,b}^{-1}\Q_{b,c}\[\u\]_c,
\end{eqnarray}
where the {\em{message}} $\M$ is defined as
\begin{eqnarray}
  \M=\Q_{c,b}\Q_{b,b}^{-1}\Q_{b,c}.
\end{eqnarray}
Otherwise, if $\Q_{b,b}$ is singular then the solution is simply
\begin{eqnarray}
\label{ac0}  \[\u\]_{C_1}&=&\0\\
  \[\u\]_b&=&\u_{\rm{null}}\(\K_{b,b}\).
\end{eqnarray}
This singular case is highly unlikely as the probability of (\ref{ac0}) in continuous models is zero. However, it should be checked for completeness.

\subsection{Solution: higher order components}\label{sec_high}
In practice, dimensionality reduction involves the projection of the data into the subspace of a few of the first principal components. We now show that the algorithm in \ref{sec_eigval} can be extended to provide higher order components.

The $j$'th principal component is defined as the linear transformation which is orthogonal to the preceding components and preserves maximal variance. Similarly to the first component it is given by $\X_j=\u_j^T\x$ where $\u_j$ is the $j$'th principal eigenvector of $\S$. In the concentration domain, $\u_j$ is the eigenvector associated with $\lambda_j$, the $j$'th smallest eigenvalue of $\K$.

In order to distribute the computation of $\lambda_j$, we adjust (\ref{eigasopt}) using the following lemma:
\begin{lemma}
   Let $\K$ be a symmetric matrix with eigenvalues $\lambda_1\leq,\cdots,\leq\lambda_p$ and eigenvectors $\u_1,\cdots,\u_p$. Then,
  \begin{eqnarray}\label{suptv}
    \lambda_j=\sup_{\{v_i\}_{i=1}^{j-1},t}t\quad{\rm{s.t.}}\quad t<{\rm{eig}}_{\min}\(\K+\sum_{i=1}^{j-1}v_i\u_i\u_i^T\).
  \end{eqnarray}
  The optimal $v_i$ are any values which satisfy $v_i>\lambda_j-\lambda_i$ for $i=1,\cdots,j-1$.
\end{lemma}
\begin{IEEEproof}
The proof is based on the recursive variational characterization of of the $j$'th smallest eigenvalue\footnote{There is also a non-recursive characterization known as Courant-Fischer theorem which results in a similar maximin representation \cite{Golub:83}.}:
\begin{eqnarray}\label{dualuk}
  \lambda_j=\left\{\begin{array}{ll}
                    \min_{\u} & \u^T\K\u \\
                    {\rm{s.t.}} & \u^T\u=1\\
                    & \u^T\u_i=0, \quad i=1,\cdots,j-1
                  \end{array}
  \right.
\end{eqnarray}
where $\u_i$ are the preceding eigenvectors, and the optimal solution $\u_j=\u$ is the eigenvector associated with $\lambda_j$. A dual representation can be obtained using Lagrange duality. We rewrite the orthogonality restrictions as quadratic constraints $\u^T\u_i\u_i^T\u=0$ and eliminate them using Lagrange multipliers:
\begin{eqnarray}
  \lambda_j\geq\max_{t,\{v_i\}_{i=1}^{j-1}}\min_{\u} t+\u^T\[\K-t\I+\sum_{i=1}^{j-1}v_i\u_i\u_i^T\]\u
\end{eqnarray}
where the inequality is due to the weak duality \cite{boyd:2003}. The inner minimization is unbounded unless
\begin{eqnarray}
  \K-t\I+\sum_{i=1}^{j-1}v_i\u_i\u_i^T\succeq \0.
\end{eqnarray}
Therefore,
\begin{eqnarray}\label{duallambdak}
  \lambda_j\geq\max_{t,\{v_i\}_{i=1}^{j-1}}\quad{\rm{s.t.}}\quad t\leq{\rm{eig}}_{\min}\(\K+\sum_{i=1}^{j-1}v_i\u_i\u_i^T\)
\end{eqnarray}
Lagrange duality does not guaranty an equality in (\ref{duallambdak}) since (\ref{dualuk}) is not convex. However, it is easy to see that the inequality is tight and can be attained by choosing $\u=\u_j$. Finally, (\ref{suptv}) is obtained by replacing the maximum with a supremum and relaxing the constraint.
\end{IEEEproof}

Lemma 2 allows us to find $\lambda_j$ in a distributed manner. We replace $\K$ with $\overline\K=\K+\U\D\U^T$ where $\U$ is a $p\times (j-1)$ matrix with the preceding eigenvectors as its columns and $\D$ is a $(j-1)\times(j-1)$ diagonal matrix with sufficiently high constants on its diagonal, and search for its principal component. The matrix $\overline\K$ does not necessarily satisfy the sparse block structure of $\K$ so we cannot use the solution in Proposition 1 directly. Fortunately, it can be easily adjusted since the modification to $\K$ is of low rank.
\begin{proposition}
  Let $\K$ be a symmetric matrix with $\K_{a,b}=\K_{b,a}^T=\0$. Then, the constraint
  \begin{eqnarray}\label{eigasopt2}
    t<{\rm{eig}}_{\min}\(\[\begin{array}{ccc}
    \K_{a,a} & \K_{a,c} & \0 \\
    \K_{c,a} & \K_{c,c} & \K_{c,b} \\
    \0 & \K_{b,c} & \K_{b,b}
  \end{array}
 \]+\U\D\U^T\)
  \end{eqnarray}
  is equivalent to the following pair of constraints
  \begin{eqnarray}
\label{prop2u}    t&<&{\rm{eig}}_{\min}\(\K_{b,b}+\[\U\]_{b,:}\D\[\U\]_{b,:}^T\)\\
\label{prop1u}    t&<&{\rm{eig}}_{\min}\(\K_{C_1,C_1}+\[\overline\U\]_{C_1,:}\overline\D\[\overline\U\]_{C_1,:}^T\)
  \end{eqnarray}
  where
  \begin{eqnarray}
    \[\overline\U\]_{C_1,:}&=&\[\begin{array}{cc}
           \begin{array}{c}
             \0 \\
             \I
           \end{array}
            & \[\U\]_{C_1} \\
         \end{array}
    \]\\
    \overline{\D}&=&\[\begin{array}{cc}
                      \0 & \0 \\
                      \0 & \D
                    \end{array}
    \]-\M_\U\(t\)
  \end{eqnarray}
   and the {\em{message matrix}} $\M_\U\(t\)$ is defined as
  \begin{eqnarray}
    \label{messageU}\M_\U\(t\)&=&\[\begin{array}{c}
                               \K_{c,b} \\
                               \D\[\U\]_{b,:}^T
                             \end{array}
    \]\(\K_{b,b}+\[\U\]_{b,:}\D\[\U\]_{b,:}^T-t\I\)^{-1}\[\begin{array}{cc}
                               \K_{b,c} & \[\U\]_{b,:}\D
                             \end{array}
    \].
  \end{eqnarray}
\end{proposition}
\begin{IEEEproof}
  The proof is similar to that of Proposition 1 and therefore omitted.
\end{IEEEproof}

Thus, the solution to the $j$'th largest eigenvalue is similar to the method in Section \ref{sec_eigval}. The only difference is that the messages are slightly larger. Each message is a matrix of size $|S|+j-1\times |S|+j-1$. In practice, dimensionality reduction involves only a few principal components and this method is efficient when $|S|+j-1$ is considerably less than $p$ (the size of the messages in a fully centralized protocol).

The higher order components can therefore be found in a distributed manner as detailed in Section \ref{sec_eigvec} above.

\section{DPCA in decomposable graphs}\label{sec_dpca}
We now proceed to the general problem of DPCA in decomposable graphs. In the previous section, we showed that DPCA can be computed in a distributed manner if it is a priori known that $\x_a$ and $\x_b$ are conditionally independent given $\x_c$. Graphical models are intuitive characterizations of such conditional independence structures. In particular, decomposable models are graphs that can be recursively subdivided into the two cliques graph in Fig. 1. Therefore, this section consists of numerous technical definitions taken from \cite{lauritzen:book} followed by a recursive application of the previous results.

An undirected graph $\mathcal{G}$ is a set of nodes connected by undirected edges. A random vector $\x$ satisfies the Markov property with respect to $\mathcal{G}$, if for any pair of non-adjacent nodes the corresponding pair of random variables are conditionally independent on the rest of the elements in $\x$. In the Gaussian distribution, this definition results in sparsity in the concentration domain. If $\K$ is the concentration matrix of a jointly Gaussian multivariate $\x$ that satisfies $\mathcal{G}$, then $\[\K\]_{i,j}=0$ for any pair $\{i,j\}$ of non-adjacent nodes.

Decomposable graphs are a specific type of graph which possess an appealing structure. A graph is decomposable if it can be recursively be subdivided into disjoint sets of nodes $a$, $b$ and $c$, where $c$ separates $a$ and $b$, and $c$ is complete, i.e., there are no edges between $a$ and $b$ and all the nodes within $c$ are connected by an edge. Clearly, the simplest non-trivial decomposable graph is the two cliques graph in Fig. \ref{fig_twoclique}.

A clique is a maximal subset of nodes which is fully connected. It is convenient to represent a decomposable graph using a sequence of cliques $C_1,\cdots,C_K$ which satisfy a {\em{perfect elimination order}}. An important property of this order is that $S_j$ separates
$H_{j-1}\backslash S_j$ from $R_j$ where
\begin{eqnarray}\label{HSR}
  H_{j}&=&C_1\cup C_2\cup\cdots\cup C_{j},\quad j=1,\cdots,K\\
  S_{j}&=&H_{j-1}\cap C_j, \quad j=2,\cdots,K\\
  R_{j}&=&H_j\backslash H_{j-1}, \quad j=2,\cdots,K.
\end{eqnarray}
Note that this perfect elimination order induces an inherent asymmetry between the cliques which will be used in our recursive solution below. The two cliques graph in Fig. \ref{fig_twoclique} is a simple example of a decomposable graph with $C_1=\{a,c\}$, $C_2=\{c,b\}$, $S_2=\{c\}$, $H_1=\{a,c\}$, $H_2=\{a,c,b\}$ and $R_2=\{b\}$. Accordingly, $S_2=\{c\}$ separates $H_1\backslash S_2=\{a\}$ from $C_2\backslash S_2=\{b\}$.

Similarly to the previous section, global ML estimation of the concentration matrix in decomposable Gaussian graphical model has a simple closed form. It can be computed in a distributed manner:
\begin{eqnarray}\label{decomp_ml}
  \K=\sum_{k=1}^K\[\tilde\K^{C_k,C_k}\]^0-\sum_{k=2}^K\[\tilde\K^{S_k,S_k}\]^0
\end{eqnarray}
where the local estimates are defined as:
\begin{eqnarray}
\label{decomp_ml1}  \tilde\K^{C_k,C_k}=\(\tilde\S^{C_k,C_k}\)^{-1},\quad k=1,\cdots,K\\
\label{decomp_ml2}  \tilde\K^{S_k,S_k}=\(\tilde\S^{S_k,S_k}\)^{-1},\quad k=2,\cdots,K,
\end{eqnarray}
and
\begin{eqnarray}
\label{decomp_ml3}  \tilde\S^{C_k,C_k}=\frac{1}{n}\sum_{i=1}^n\[\x_i\]_{C_k}\[\x_i\]_{C_k}^T,\quad k=1,\cdots,K\\
\label{decomp_ml4}  \tilde\S^{S_k,S_k}=\frac{1}{n}\sum_{i=1}^n\[\x_i\]_{S_k}\[\x_i\]_{S_k}^T,\quad k=2,\cdots,K.
\end{eqnarray}
The zero fill-in operator $\[\cdot\]^0$ in (\ref{decomp_ml}) outputs a matrix of the same dimension as $\K$ where the argument occupies the appropriate sub-block and the rest of the matrix has zero valued elements (See (\ref{Kabc}) for a two clique example, and \cite{lauritzen:book} for the exact definition of this operator).

DPCA can be recursively implemented by using the previous two clique solution. Indeed, Proposition 1 shows that the eigenvalue inequality
\begin{eqnarray}
  t<{\rm{eig}}_{\min}\(\K\)
\end{eqnarray}
is equivalent to two adjusted local eigenvalue inequalities
\begin{eqnarray}
  t&<&{\rm{eig}}_{\min}\(\K_{R_K}'\(t\)\)\\
  \label{eigminHk1}t&<&{\rm{eig}}_{\min}\(\K_{H_{K-1}}'\(t\)\)
\end{eqnarray}
where
\begin{eqnarray}
  \K_{R_K}'\(t\)&=&\K_{R_K,R_K}\\
  \K_{H_{K-1}}'\(t\)&=&\K_{H_{K-1},H_{K-1}}\(t\)-\[\M_k\(t\)\]^0.
\end{eqnarray}
where $\M_k\(t\)$ is a message as in (\ref{message}) and $\[\cdot\]^0$ is the zero fill-in operator. Next, we can apply Schur's Lemma again and replace (\ref{eigminHk1}) with two additional inequalities:
\begin{eqnarray}
  t&<&{\rm{eig}}_{\min}\(\K_{R_K}'\(t\)\)\\
  t&<&{\rm{eig}}_{\min}\(\K_{R_{K-1}}''\(t\)\)\\
  t&<&{\rm{eig}}_{\min}\(\K_{H_{K-2}}''\(t\)\)
\end{eqnarray}
where $\K_{R_{K-1}}''\(t\)$ and $\K_{H_{K-2}}''\(t\)$ are similarly defined. We continue in an iterative fashion until we obtain $K$ decoupled eigenvalue inequalities. Thus, the feasibility of a given $t$ can be checked in a distributed manner with minimal message passing between the cliques, and any line-search can efficiently solve DPCA.

Specifically, in Algorithm 1 displayed below we provide a pseudo code for DPCA that solves for $t$ using the bisection method. Given initial bounds
\begin{eqnarray}
  L\leq {\rm{eig}}_{\min}\(\K\)\leq U,
\end{eqnarray}
Algorithm 1 is guaranteed to find the minimal eigenvalue up to any required tolerance $\epsilon$ within $\log_2\frac{U-L}{\epsilon}$ iterations. Each iteration consists of up to $K-1$ messages through the matrices $\M_k\(t\)$ whose dimensions are equal to the cardinalities of $S_k$ for $k=2,\cdots,K$. A simple choice for the bounds is $L=0$ since $\K$ is positive definite, and
\begin{eqnarray}
  U=\min_{k=1,\cdots,K}\{\rm{eig}_{\min}\(\K_{C_k,C_k}\)\}
\end{eqnarray}
as proved in the Appendix.

\begin{algorithm}
\dontprintsemicolon
\SetLine
\KwIn{$\K$, $L$, $U$, $\epsilon$, clique tree structure}
\KwOut{$t$}
\While{$U-L>\epsilon$}{
$t=\(U+L\)/2$\;
$\Q=\K$\;
\For{$k=K,\cdots,2$}{
\eIf{$t<{\rm{eig}}_{\min}\(\Q_{R_k,R_k}\)$}
{$\M_k\(t\)=\Q_{S_k,R_k}\(\Q_{R_k,R_k}-t\I\)^{-1}\Q_{R_k,S_k}$ \;$\Q_{S_k,S_k}=\Q_{S_k,S_k}-\M_k\(t\)$\;}
{$U= t$\; \bf{break loop}}
}
\If{$U>t$}
{\eIf{$t<{\rm{eig}}_{\min}\(\Q_{C_1,C_1}\)$}{$L=t$\;}{$U=t$\;}}
}
\caption{Bisection line search for DPCA}
\label{bisection}
\end{algorithm}

Given a principal eigenvalue $\lambda$, its corresponding eigenvector can be computed by solving $\Q\u=\0$ where $\Q=\K-\lambda\I$ as detailed in Section \ref{sec_eigvec}. Beginning with $k=K$ we partition $H_{k}$ into $R_k$ and $H_{k-1}$ and test the singularity of $\Q_{R_k,R_k}$. If it is singular, then $\lambda$ is associated with $R_k$. Otherwise, we send the message $\M_k\(\lambda\)$ to $\H_{k-1}$ and repartition it. We continue until we find the associated remainder $R_k$ or reach the first clique. Then, we compute the corresponding local null vector and begin propagating it to the higher remainders as expressed in (\ref{ub}). A pseudo code of this method is provided in Algorithm 2 below.

\begin{algorithm}
\dontprintsemicolon
\SetLine
\KwIn{$\Q$, clique tree structure}
\KwOut{$\u$}
$\u=\0$\;
$\Q=\K$\;
$k=K$\;
\While{$\(k>1\)\&\(\Q_{R_k,R_k}{\rm{\;non\;singular}}\)$}
{$\M_k=\Q_{S_k,R_k}\Q_{R_k,R_k}^{-1}\Q_{R_k,S_k}$\; $\Q_{S_k,S_k}=\Q_{S_k,S_k}-\M_k$\;$k=k-1$\;}
$\u\(C_k\)=\u_{\rm{null}}\(\Q_{C_k,C_k}\)$\;
\For{$k=k+1,\cdots,K$}
{$\u\(R_k\)=-\Q_{R_k,R_k}^{-1}\Q_{R_k,S_k}\u\(S_k\)$\;}
\caption{Eigenvector computation via $\Q\u=\0$}
\label{eigvectorcode}
\end{algorithm}

Algorithm 1 can be easily extended to compute higher order eigenvalues through application of Proposition 2. For this purpose, note that the inequality in (\ref{prop1u}) has the same structure as (\ref{eigasopt2}) and therefore can be recursively partitioned again. The only difference is that the rank of the modification is increased at each clique and requires larger message matrices. Thus, the algorithm is efficient as long as the size of the separators ($|S_k|$), the number of cliques ($K$) and the number of required eigenvalues ($j$) are all relatively small in comparison to $p$. Given any eigenvalue (first or high order), Algorithm 2 finds the associated eigenvector in a distributed and efficient manner.

\section{Synthetic tracking example}\label{sec_num}
\begin{figure}\center
\includegraphics[width=0.40\textwidth]{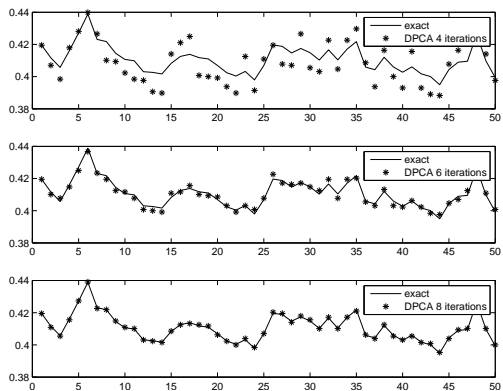}
\caption{Iterations of the DPCA bisection line-search in a time varying scenario.
\label{timevarying}}
\end{figure}

We now illustrate the performance of DPCA using a synthetic numerical example. Specifically, we use DPCA to track the first principle component in a slowly time varying setting. We define a simple graphical model with $305$ nodes representing three fully connected networks with only 5 coupling nodes, i.e., $C_1=\{1,\cdots,100,301,\cdots,305\}$, $C_2=\{101,\cdots,200,301,\cdots,305\}$, and $C_3=\{201,\cdots,300,301,\cdots,305\}$. We generate $5500$ length $p=305$ vectors $\x_i$ of zero mean, unit variance and independent Gaussian random variables. At each time point, we define $\K$ through (\ref{decomp_ml})-(\ref{decomp_ml4}) using a sliding window of $n=500$ realizations with $400$ samples overlap. Next, we run DPCA using Algorithm 1. Due to slow time variation, we define the lower ($L$) and upper ($U$) bounds as the value of the previous time point minus and plus 0.1, respectively. We define the tolerance as $\epsilon=0.001$ corresponding to 8 iterations.  Figure \ref{timevarying} shows the exact value of the minimal eigenvalue as a function of time along with its DPCA estimates at the $4$'th, $6$'th and $8$'th iterations. It is easy to see that a few iterations suffice for tracking the maximal eigenvalue at high accuracy. Each iteration involves three EVDs of approximately $105\times 105$ matrices and communication through two messages of size $5\times 5$. For comparison, a centralized solution would require sending a set of $100$ length $305$ vectors to a central processing unit which computes an EVD of a matrix of size $305\times 305$.

\section{Application to distributed anomaly detection in networks}\label{sec_abilene}
A promising application for DPCA is distributed anomaly detection in computer networks. In this context, PCA is used for learning a low dimensional model for normal behavior of the traffic in the network. The samples are projected into the subspace associated with the first principal components. Anomalies are then easily detected by examining the residual norm. Our hypothesis is that the connectivity map of the network is related to its statistical graphical model. The intuition is that two distant links in the network are (approximately) independent conditioned on the links connecting them and therefore define a graphical model. We do not rigorously support this claim but rather apply it in a heuristic manner in order to illustrate DPCA.

\begin{figure}\center
\includegraphics[width=0.40\textwidth]{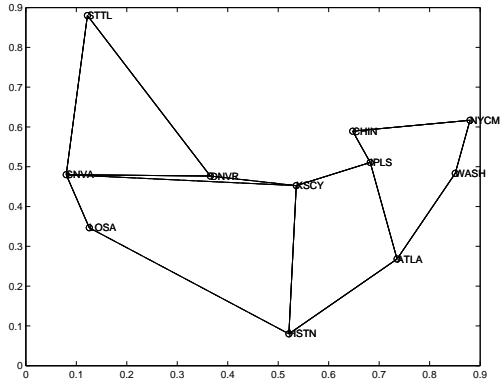}
\caption{Map of the Abilene network.
\label{abilene_map}}
\end{figure}

Following  \cite{Lakhina:2004,huang:nips2006}, we consider a real world dataset of Abilene, the Internet2 backbone network. This network carries traffic from universities in the United States. Figure \ref{abilene_map} shows its connectivity map consisting of 11 routers and 41 links (each edge corresponds to two links and there are additional links from each of the nodes to itself). Examining the network it is easy to see that the links on the east and west sides of the map are separated through six coupling links: DNVR-KSCY, SNVA-KSCY and LOSA-HSTN. Thus, our first approximated decomposable graph, denoted by $\mathcal{G}_{\rm{2 cliques}}$, consists of two cliques: an eastern clique and a western clique coupled by these six links. Graph $\mathcal{G}_{\rm{2 cliques}}$ corresponds to a decomposable concentration matrix with a sparsity level of $0.33$. Our second decomposable graph denoted by $\mathcal{G}_{\rm{3 cliques}}$ is obtained by redividing the eastern clique again into two cliques separated through the four coupling links: IPLS-CHIN and ATLA-WASH. Its corresponding concentration matrix has a sparsity level of $0.43$. Finally, for comparison we randomly generate an arbitrary graph $\mathcal{G}_{\rm{random}}$ over the Abilene nodes, with an identical structure as $\mathcal{G}_{\rm{3 cliques}}$ (three cliques of the same cardinalities), which is not associated with the topology of the Abilene network.

\begin{figure}\center
\includegraphics[width=0.40\textwidth]{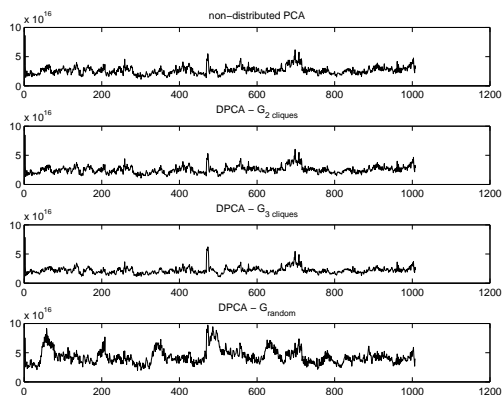}
\caption{Projection into anomaly subspace with and without graphical models.
\label{abilene_fig1}}
\end{figure}

\begin{figure}\center
\includegraphics[width=0.40\textwidth]{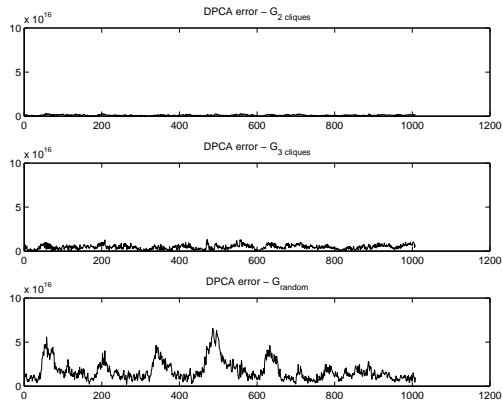}
\caption{Absolute error in projection into anomaly subspace with different graphical models.
\label{abilene_fig2}}
\end{figure}

In our experiments, we learn the $41\times 41$ covariance matrix from a $41\times 1008$ data matrix representing 1008 samples of the load on each of the 41 Abilene links during April 7-13, 2003. We compute PCA and project each of the $1008$ samples of dimension $41$ into the null space of the first four principal components. The norm of these residual samples is plotted in the top plot of Fig. \ref{abilene_fig1}. It is easy to see the spikes putatively associated with anomalies. Next, we examine the residuals using DPCA with $\mathcal{G}_{\rm{2 cliques}}$, $\mathcal{G}_{\rm{3 cliques}}$ and $\mathcal{G}_{\rm{random}}$. The norms of the residuals are plotted in the three lower plots of Fig. \ref{abilene_fig1}., respectively.  As expected,  the topology based plots are quite similar with spikes occurring at the times of these anomalies. Thus, we conclude that the decomposable graphical model for Abilene is a good approximation and does not cause substantial loss of information (at least for the purpose of anomaly detection). On the other hand, the residual norm using the random graph is a poor approximation as it does not preserve the anomalies detected by the full non-distributed PCA. These conclusions are supported in Fig. \ref{abilene_fig2} where we show the absolute errors of DPCA with respect to PCA using the different graphical models. It is easy to see that $\mathcal{G}_{\rm{2 cliques}}$ results in minimal error, $\mathcal{G}_{\rm{3 cliques}}$ provides a reasonable tradeoff between performance and computational complexity (through its increased sparsity level), while graph $\mathcal{G}_{\rm{random}}$ is clearly a mismatched graphical model and results in significant increase in error.

\section{Discussion and future work}\label{sec_conc}
In this paper, we introduced DPCA and derived a decentralized method for its computation. We proposed distributed anomaly detection in communication networks as a motivating application for DPCA and investigated possible graphical models for such settings.

Future work should examine the statistical properties of DPCA. From a statistical perspective, DPCA is an extension of classical PCA to incorporate additional prior information. Thus, it would be interesting to analyze the distribution of its components and quantify their significance, both under the true graphical model and under mismatched models. In addition, DPCA is based on the intimate relation between the inverse covariance and the conditional Gaussian distribution. Therefore, it is also important to assess its sensitivity to non-Gaussian sources. Finally, alternative methods to ML in singular and ill conditioned scenarios should be considered.

\section{Acknowledgments}
The authors would like to thank Clayton Scott for providing the Abilene data, and Arnau Tibau Puig for stimulating discussions and his help with the Abilene dataset.

\appendix
In this appendix, we prove that the minimal eigenvalue of a $p\times p$ symmetric matrix $\K$ is less than or equal to the minimal eigenvalue of any of its sub-matrices, say $\K_{a,a}$ for some set of indices $a$. For simplicity, we assume that $a=\{1,\cdots,p_a\}$ for some integer $p_a\leq p$. The proof is a simple application of the Rayleigh quotient characterization of the minimal eigenvalues:
\begin{eqnarray}
  {\rm{eig}}_{\min}\(\K\)&=&\min_\u\frac{\u^T\K\u}{\u^T\u}\\
  &\leq& \frac{\[\begin{array}{cc}
                  \v^T & \0^T
                \end{array}
  \]\K\[\begin{array}{c}
          \v \\
          \0
        \end{array}
  \]}{\[\begin{array}{cc}
                  \v^T & \0^T
                \end{array}
  \]\[\begin{array}{c}
          \v \\
          \0
        \end{array}
  \]}\\
  &=&\frac{\v^T\K_{a,a}\v}{\v^T\v}\\
\label{mineigv}  &=&\min_\u\frac{\u^T\K_{a,a}\u}{\u^T\u}\\
  &=&{\rm{eig}}_{\min}\{\K_{a,a}\}
\end{eqnarray}
where $\v$ is the optimal solution to (\ref{mineigv}).


\begin{thebibliography}{10}

\bibitem{Anderson:book}
T.~W. Anderson.
\newblock {\em An introduction to multivariate statistical analysis}.
\newblock John Wiley and Sons, second edition edition, 1971.

\bibitem{bai:book2005}
Z.~J. Bai, R.~H. Chan, and F.~T. Luk.
\newblock {\em Advanced Parallel Processing Technologies}, chapter Principal
  Component Analysis for Distributed Data Sets with Updating, pages 471--483.
\newblock 2005.

\bibitem{banerjee-2007}
O.~Banerjee, L.~El Ghaoui, and A.~d'Aspremont.
\newblock Model selection through sparse maximum likelihood estimation.
\newblock {\em Journal of Machine Learning Research}, 9:485--516, March 2008.

\bibitem{boyd:2003}
S.~Boyd and L.~Vandenberghe.
\newblock {\em Introduction to Convex Optimization with Engineering
  Applications}.
\newblock Stanford, 2003.

\bibitem{willsky:SPM}
M.~Cetin, L.~Chen, J.~W. Fisher, A.~T. Ihler, R.~L. Moses, M.~J. Wainwright,
  and A.~S. Willsky.
\newblock Distributed fusion in sensor networks: A graphical models
  perspective.
\newblock {\em IEEE Signal Processing Magazine}, 23(4):42-- 55, July 2006.

\bibitem{Chhabra:2008}
P.~Chhabra, C.~Scott, E.~Kolaczyk, and M.~Crovella.
\newblock Distributed spatial anomaly detection.
\newblock In {\em Proceedings of INFOCOM}, April 2008.

\bibitem{dempster:72}
A.~P. Dempster.
\newblock Covariance selection.
\newblock {\em Biometrics}, 28:157--175, 1972.

\bibitem{friedman-2007}
J.~Friedman, T.~Hastie, and R.~Tibshirani.
\newblock Sparse inverse covariance estimation with the {L}{A}{S}{S}{O}.
\newblock {\em Biostat}, 9(3):432 -- 441, July 2008.

\bibitem{Gastpar:06}
M.~Gastpar, P.~L. Dragotti, and M.~Vetterli.
\newblock The distributed {K}arhunen {L}oeve transform.
\newblock {\em IEEE Trans. on Information Theory}, 52(12):5177--5196, Dec.
  2006.

\bibitem{Golub:83}
G.~H. Golub and C.~F.~Van Loan.
\newblock {\em Matrix Computations}.
\newblock John Hopkins, 1983.

\bibitem{hastietf01}
T.~Hastie, R.~Tibshirani, and J.~Friedman.
\newblock {\em The elements of statistical learning: {D}ata mining, inference,
  and prediction}.
\newblock Springer, New York, 2001.

\bibitem{huang:nips2006}
L.~Huang, X.~Nguyen, M.~Garofalakis, M.~I. Jordan, A.~D. Joseph, and N.~Taft.
\newblock In-network {P}{C}{A} and anomaly detection.
\newblock In {\em Proceedings of NIPS'2006}, Dec. 2006.

\bibitem{jordan:book}
M.~I. Jordan.
\newblock {\em Introduction to graphical models}.
\newblock Unpublished, 2008.

\bibitem{Kargupta:2001}
H.~Kargupta, W.~Huang, K.~Sivakumar, and E.~Hohnson.
\newblock Distributed clustering using collective principal component analysis.
\newblock {\em Knowledge and Information Systems}, 3(4):422--448, Nov. 2001.

\bibitem{Lakhina:2004}
Anukool Lakhina, Mark Crovella, and Christophe Diot.
\newblock Diagnosing network-wide traffic anomalies.
\newblock {\em SIGCOMM Comput. Commun. Rev.}, 34(4):219--230, 2004.

\bibitem{lauritzen:book}
S.~L. Lauritzen.
\newblock {\em Graphical models}, volume~17.
\newblock Oxford Statistical Science Series, New York, 1996.

\bibitem{qu:icdm2002}
Y.~Qu, G.~Ostrouchovz, N.~Samatovaz, and A.~Geist.
\newblock Principal component analysis for dimensions reduction in massive
  distributed data sets.
\newblock In {\em IEEE International Conference on Data Mining (ICDM)}, 2002.

\bibitem{roy:08}
O.~Roy and M.~Vetterli.
\newblock Dimensionality reduction for distributed estimation in the infinite
  dimensional regime.
\newblock {\em IEEE Trans. on Information Theory}, 54(2):1655--1669, April
  2008.

\bibitem{Schizas:2007}
I.~D. Schizas, G.~B. Giannakis, and Z.~Q. Luo.
\newblock Distributed estimation using reduced-dimensionality sensor
  observations.
\newblock {\em IEEE Trans. on Signal Processing}, 55(8):4284--4299, Aug. 2007.

\bibitem{YairWeiss10012001}
Y.~Weiss and W.~T. Freeman.
\newblock {Correctness of Belief Propagation in Gaussian Graphical Models of
  Arbitrary Topology}.
\newblock {\em Neural Comp.}, 13(10):2173--2200, 2001.

\bibitem{xiao:2006}
J.~J. Xiao, A.~Ribeiro, Z.~Q. Luo, and G.~B. Giannakis.
\newblock Distributed compression-estimation using wireless sensor networks.
\newblock {\em Signal Processing Magazine}, 23(4):27-- 41, July 2006.

\bibitem{yuan:2007}
M.~Yuan and Y.~Lin.
\newblock Model selection and estimation in the gaussian graphical model.
\newblock {\em Biometrika}, 94(1):19--35, 2007.

\bibitem{zhu:may2005}
Y.~Zhu, E.~Song, J.~Zhou, and Z.~You.
\newblock Optimal dimensionality reduction of sensor data in multisensor
  estimation fusion.
\newblock {\em IEEE Trans. on Signal Processing}, 53(5):1631--1639, May 2005.

\end{thebibliography}
\end{document}